\documentclass[10pt,logo,copyright]{nvidiatechreport}
\usepackage{parskip}
\usepackage{natbib}
\setcitestyle{authoryear,round,citesep={;},aysep={,},yysep={;}}
\renewcommand{\today}{}
\makeatletter
\renewcommand{\author}[1]{%
  \gdef\@author{{\Authfont\shortstack[l]{#1}}}%
}
\makeatother
\usepackage[T1]{fontenc}
\usepackage{amsmath,amssymb,amsthm,booktabs,graphicx}
\usepackage{algorithm,algpseudocode}
\usepackage{microtype}
\usepackage{xcolor}
\usepackage{colortbl}
\usepackage{pgfplots}
\pgfplotsset{compat=1.18}
\usepackage{hyperref}
\usepackage{url}

\definecolor{rpgablue}{RGB}{28,83,132}
\definecolor{denseorange}{RGB}{179,87,27}
\hypersetup{colorlinks=true,allcolors=nvidiagreen}

\newcommand{\methodname}{LoGRA}

\title{\methodname{}: Scaling LLM Reinforcement Learning with Low-Rank Gradient Sketches}
\author{%
Shaokun Zhang, Yifan Zhang, Jian Hu, Yueying Li,
Hao Zhang, Binfeng Xu, Jan Kautz, Yi Dong  \\[6pt]
NVIDIA
}
\hypersetup{pdfauthor={Shaokun Zhang, Yifan Zhang, Jian Hu, Yueying Li, Hao Zhang, Binfeng Xu, Jan Kautz, Yi Dong}}

\begin{document}
\maketitle

\begin{abstract}
 Reinforcement learning has greatly advanced the capabilities of large language models, but its memory demands remain a barrier to broader adoption. 
 We introduce \methodname{}, an approach to RL post-training that reduces memory by retaining useful learning signals in low-rank gradient sketches. These compact representations support both model updates and efficient policy synchronization. To prevent overly large updates from disrupting learning, we complement gradient compression with predicted-KL step control, which estimates policy changes before applying each update and adjusts its magnitude accordingly. 
 With all techniques combined, \methodname{} reduces average training memory usage by up to 45.7\% across reasoning tasks without compromising performance.
 It also enables stable training of a 27B-parameter model for over 1,100 steps on a single eight-GPU node, where dense Adam runs out of memory, making previously memory-infeasible RL training practical. Code is available in the \href{https://github.com/skzhang1/labs-molt/tree/logra/examples/scripts/logra}{Molt library}.
\end{abstract}
\abscontent

\section{Introduction}

\begin{figure}[htb]
\centering
\includegraphics[width=\linewidth]{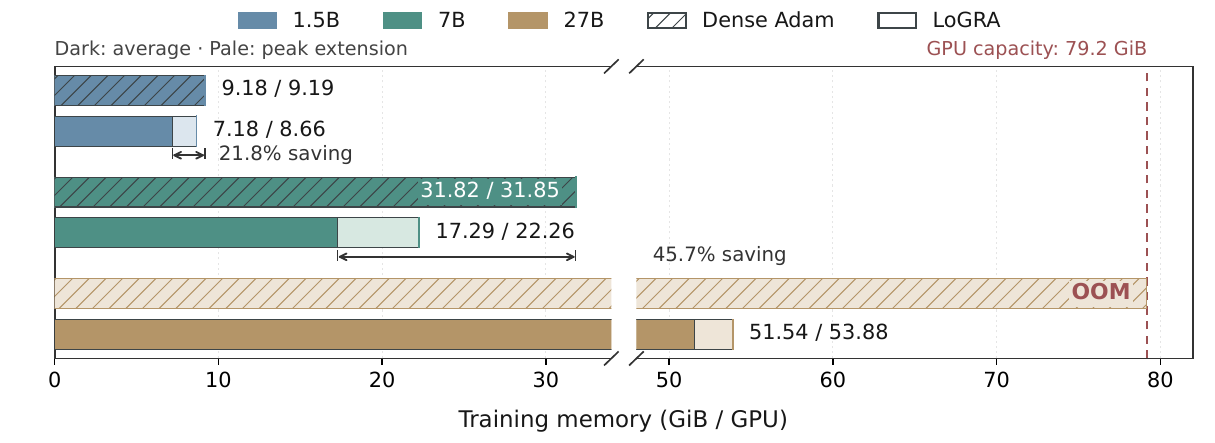}
\caption{\textbf{Lower training memory across model scales.} Colors denote model sizes; hatching distinguishes Dense Adam from LoGRA. Dark bars show average memory and pale extensions reach peak memory (Table~\ref{tab:plotted-runs-systems}); labels report average / peak in GiB per training GPU. Arrows show average-memory savings over Dense Adam. The broken axis omits 34--48 GiB. The 27B Dense Adam bar marks OOM at GPU capacity, not a measured value.}
\label{fig:intro-memory}
\end{figure}

Reinforcement learning (RL) has emerged as a key driver of advances in LLM, enabling models to improve through feedback on their own generated responses~\citep{sheng2024hybridflow,nemo-rl,xu2026polar,zhang2026nemotron}. 
However, its substantial memory requirements remain a major barrier to broader adoption. This challenge extends well beyond storing model parameters, as gradients and optimizer states can consume several times more memory than the weights themselves. For example, the BF16 weights of a 7B-parameter Qwen model~\citep{bai2023qwen} require approximately 14 GB, while Adam’s~\citep{kingma2014adam} two FP32 moment buffers add another 56 GB, even before accounting for gradients and activations. Consequently, hardware capable of running a model may still be insufficient to improve it through RL. Reducing this memory overhead is therefore critical to making RL post-training accessible to the broader research community.

To overcome this bottleneck, existing approaches span both system-level optimizations and memory-efficient training algorithms.
At the system level, FSDP~\citep{zhao2023pytorch} shards training states across GPUs, gradient checkpointing~\citep{chen2016training} reduces activation storage through recomputation, and  CPU offloading moves selected training states to host memory~\citep{rajbhandari2020zero}. 
Parameter-efficient methods reduce the number of parameters that require gradients and optimizer states: LoRA~\citep{hu2021lora}, for example, freezes pretrained weights and learns small low-rank adapters. 
Gradient compression offers another route~\citep{zhao2024galore, su2025galore,hao2024flora}: they keep the original weight matrices trainable while using compact representations of gradients and optimizer states, thereby reducing training memory while still enabling effective updates to the full model.

Motivated by this perspective, we introduce \methodname{}, a method that uses low-rank gradient sketches to reduce the memory and communication costs of RL post-training, thus lower the hardware barrier. 
Our intuition is that learning from rewards may require far less information than a full gradient can represent: with binary outcome rewards, each generated response receives only a single correctness signal, motivating a compact representation of the resulting learning signal~\citep{schulman2025lora}. 
\methodname{} directly accumulates sketches during backpropagation, avoiding the construction and storage of full gradients for the selected weight matrices. These sketches produce updates that are merged into the model weights, with directions that can vary throughout training. The same compact representation is then reused to synchronize the rollout policy, extending the benefits of compression from training memory to communication.

LoGRA alone is insufficient to ensure stable updates: an overly large update can abruptly shift the policy and disrupt previously learned behavior. We therefore introduce predicted-KL step control, which estimates how much a proposed update would change the model’s next-token probabilities before applying it. This estimate guides the update scale to stay within a prescribed KL budget with simple optimizers.

We evaluate \methodname{} across multiple reasoning tasks with models ranging from 1.5B to 27B parameters. \methodname{} reduces average training memory by 21.8\% at 1.5B and 45.7\% at 7B without sacrificing task performance. These savings also make larger-scale training feasible: \methodname{} sustains stable learning of a 27B model for over 1,100 steps on a single eight-GPU node, where the dense Adam baseline runs out of memory under the same hardware allocation.
Ablations further clarify how projection design and KL budgets affect learning performance and training stability.

\section{Method}
\label{sec:rpga}

An LLM contains large weight matrices, and storing a full gradient for each matrix adds substantial training memory.
\methodname{} replaces these full gradients with compact sketches and represents each weight update as the product of two thin matrices.
We first describe this matrix representation, then explain its use in RL post-training and policy synchronization.
We use predicted-KL step control to determine the update magnitude.

\subsection{Low-rank gradient sketches}

\paragraph{Full-gradient updates.}
Let $\theta$ denote the parameters of an LLM policy $\pi_\theta$, and let $\mathcal L(\theta)$ be a differentiable RL training loss.
Consider one weight matrix $W\in\mathbb R^{d\times k}$ in the model.
Its gradient $G$ has the same dimensions as $W$; for example, an SGD update takes the form
\begin{equation}
 G=\nabla_W\mathcal L\in\mathbb R^{d\times k},
 \qquad W\leftarrow W-\eta G,
 \label{eq:dense-gradient}
\end{equation}
where $\eta$ is the learning rate.
When a training batch is split into microbatches, standard training accumulates their contributions in a full $d\times k$ gradient buffer before updating $W$.
This requires $dk$ stored values in addition to the weights and any optimizer state.
Our goal is to perform the update using a much smaller gradient representation.

\paragraph{A compact representation.}
Choose a rank $r\ll\min(d,k)$ and a random projection matrix $A\in\mathbb R^{r\times k}$.
\methodname{} represents the gradient by a sketch
\begin{equation}
 S=GA^\top\in\mathbb R^{d\times r}.
 \label{eq:sketch}
\end{equation}
Multiplication by $A^\top$ compresses the $k$ columns of $G$ into $r$ columns of $S$.
The sketch therefore stores $dr$ values instead of $dk$.
The projection $A$ is sampled independently of the current batch and held fixed throughout gradient accumulation; it is not learned.
By default, its entries are independent random signs scaled by $1/\sqrt r$.

Equation~\ref{eq:sketch} defines the sketch mathematically, but does not require computing and then compressing a full gradient.
Our implementation directly accumulates the projected gradient contributions from each microbatch into $S$ in fp32.
It does so by projecting the inputs used in the weight-gradient calculation before multiplying them by the backward derivatives.
Thus, the compressed weights do not require a full gradient buffer, while the model's ordinary forward computation is preserved.

\paragraph{A low-rank weight update.}
To update $W$, we need a matrix with the same dimensions as $W$.
The sketch $S$ has only $r$ columns, so we multiply it by $A$ to obtain a $d\times k$ approximation to the full gradient:
\begin{equation}
 \widehat G=SA\in\mathbb R^{d\times k},
 \qquad W\leftarrow W-\eta SA.
 \label{eq:logra-reconstruction}
\end{equation}
In other words, we use $SA$ in place of $G$ in the standard gradient update.
Although $SA$ has the same dimensions as $G$, it is represented by two small matrices: $S$ of size $d\times r$ and $A$ of size $r\times k$.
Because $S$ has only $r$ columns, every column of $SA$ is a combination of these $r$ columns; this is why the update has rank at most $r$.
This compact representation approximates the full gradient rather than recovering it exactly.
The resulting update is added directly to $W$, which remains a full-sized weight matrix.

An optimizer can also adjust the sketch before it is used in the update.
We call this adjusted sketch $U$; it has the same dimensions as $S$, and $U=S$ for the basic SGD variant.
The update then becomes
\begin{equation}
 W\leftarrow W-\alpha\eta UA.
 \label{eq:target}
\end{equation}
Here $\eta$ sets the nominal step size, and $\alpha$ is an additional multiplier: $\alpha=1$ leaves the step unchanged, while a smaller value reduces it.
The predicted-KL controller described below chooses $\alpha$ without changing the shapes of $U$ and $A$.

\subsection{Applying \methodname{} to RL post-training}
\label{sec:systems}

We apply the sketch to attention and MLP projection matrices, holding other parameters fixed.
Algorithm~\ref{alg:logra-training} summarizes the procedure, which supports differentiable RL losses.
Each matrix has its own $A$, $S$, and $U$, while $\alpha$ is shared.
Updates are merged directly into the weights, with no persistent adapter branch.
Refreshing $A$ each step allows successive updates to use different subspaces, so their sum need not have rank at most $r$. Here $\mathcal L_m$ denotes microbatch $m$'s contribution to $\mathcal L$.
The projected gradient in the algorithm is accumulated directly, without forming the full gradient matrix.

\paragraph{Memory and synchronization.}
The gradient accumulator shrinks from $dk$ to $dr$ values: for $d=k=4096$ and $r=64$, fp32 storage falls from 64\,MiB to 1\,MiB.
Model weights, cached projections, activations, and other buffers still consume memory.
Basic SGD requires no optimizer moments; stateful sketch optimizers add their own storage.
For synchronization, the rollout engine regenerates $A$ from its seed and applies the received factor $\alpha\eta U$.
The payload is $O(dr)$ values plus a seed and metadata, instead of a full $d\times k$ update.

\begin{algorithm}[H]
\caption{\methodname{} with projection refreshing}
\label{alg:logra-training}
\begingroup
\algtext*{EndFor}
\begin{algorithmic}[1]
\Require Policy $\pi_\theta$, selected matrices $W$, rank $r$, step size $\eta$
\State Initialize rollout weights from $\theta$.
\For{each training step}
    \State Sample responses from $\pi_\theta$; construct $\mathcal L$. \Comment{\textbf{1. Prepare}}
    \For{each $W$}
        \State Sample $A\in\mathbb R^{r\times k}$ from a fresh seed.
        \State $S\leftarrow 0\in\mathbb R^{d\times r}$
    \EndFor
    \For{each microbatch $m$ ($\theta$ and $A$ fixed)} \Comment{\textbf{2. Accumulate sketches}}
        \For{each $W$}
            \State $S\leftarrow S+(\nabla_W\mathcal L_m)A^\top$ \Comment{Direct sketch accumulation}
        \EndFor
    \EndFor
    \State For each $W$: $U\leftarrow S$ (or use a sketch optimizer). \Comment{\textbf{3. Choose the update}}
    \State $\alpha\leftarrow 1$ \Comment{Or use predicted KL; Section~\ref{sec:predictive}}
    \For{each $W$} \Comment{\textbf{4. Update and synchronize}}
        \State $W\leftarrow W-\alpha\eta UA$ \Comment{Trainer}
        \State Send $\alpha\eta U$ and the seed for $A$ to rollout.
        \State Regenerate $A$; apply the same update. \Comment{Rollout}
    \EndFor
    \State Wait for synchronization to finish.
\EndFor
\end{algorithmic}
\endgroup
\end{algorithm}

\subsection{Step-size control with predicted KL}
\label{sec:predictive}

\methodname{} specify how to change the weights, but the size of that change also matters.
An update that is small in parameter space can still substantially change the policy's output probabilities. 
Following the trust-region motivation of TRPO~\citep{schulman2015trust}, we therefore control the update magnitude using KL divergence between the current and proposed policies.

We estimate how much the proposed update would change the model's next-token probabilities, then scale the update to fit a prescribed budget.
Let $D$ denote the proposed update to all parameters; its component for each compressed weight matrix $W$ is $D_W=-\eta UA$, using the factors and nominal step size defined above.
We apply $\theta\leftarrow\theta+\alpha D$, where $\alpha$ controls the step size without changing its direction.

We select a small subset of responses from the current training batch and examine the predictions used to generate their tokens.
For each prediction, the model sees $s$ which is a prompt and the response generated so far.
Let $\mathcal P$ collect these contexts.
At each context, we compare the current model's probabilities over the vocabulary, $\pi_\theta(\cdot\mid s)$, with those of the proposed updated model, $\pi_{\theta+\alpha D}(\cdot\mid s)$.
Both are evaluated on the same text context and at the same temperature $T>0$: a higher temperature gives a flatter distribution.

KL divergence measures how much these two probability distributions differ.
We average this change over the examined predictions as derived in detail in the appendix:
\begin{equation}
 \frac{1}{|\mathcal P|}\sum_{s\in\mathcal P}
 D_{\mathrm{KL}}\!\left(\pi_\theta(\cdot\mid s)\,\middle\|\,
 \pi_{\theta+\alpha D}(\cdot\mid s)\right)
 \approx\alpha^2q(D).
 \label{eq:quadratic}
\end{equation}
The side is the average KL between the current and proposed policies over the examined contexts.
On the right, $q(D)$ is our estimate of this average change for the unscaled proposal ($\alpha=1$).
The approximation predicts a quadratic dependence on step size: halving $\alpha$ reduces the predicted KL by a factor of four.

We specify a KL budget $\delta>0$, the allowed average predicted change in next-token probabilities for one training step.
A smaller budget keeps the updated policy closer to the current one.
Equation~\ref{eq:quadratic} gives a way to control this change: rescale the proposed update by $\alpha$ so that $\alpha^2q(D)\leq\delta$.
Thus, we first estimate the proposal's predicted KL $q(D)$, then choose its scale to fit the budget.

To estimate $q(D)$, we examine how the proposal would change the scores assigned to candidate next tokens.
Let $v$ index vocabulary tokens, $z_v(s)$ be the score (logit) of token $v$ at context $s$, and $z(s)$ collect these scores.
The vector $p(s)=\operatorname{softmax}(z(s)/T)$ contains their probabilities at temperature $T$.
Let $\dot z_v(s)$ be the rate at which score $z_v(s)$ changes along $D$, evaluated at the current weights.
The predicted KL is
\begin{equation}
 q(D)=\frac{1}{2|\mathcal P|T^2}\sum_{s\in\mathcal P}
 \operatorname{Var}_{v\sim p(s)}\!\left[\dot z_v(s)\right].
 \label{eq:fisher}
\end{equation}
Derived could be found in the appendix. Here the variance measures how differently the token scores change, weighted by their current probabilities $p(s)$, and $|\mathcal P|$ is the number of examined contexts.

\paragraph{Rescale the update to fit the budget.}
Once $q(D)$ is available, the ratio $\delta/q(D)$ tells us how the budget compares with the proposed change.
Because predicted KL scales with $\alpha^2$, we take the square root of this ratio.
With a chosen upper limit $\alpha_{\max}>0$ on the update multiplier, we set
\begin{equation}
 \alpha=\min\!\left\{\alpha_{\max},\sqrt{\delta/q(D)}\right\},
 \label{eq:alpha}
\end{equation}
for finite positive $q(D)$.
For example, a proposal with $q(D)=4\delta$ is scaled to at most one half of its original size.
The cap prevents excessive enlargement when the predicted KL is small.
This rule controls the predicted change; the realized KL can differ because the estimate is a local approximation.
A constant budget keeps $\delta$ fixed, while an annealed budget gradually lowers it to permit less policy change later in training.

\section{Experiments}

We conduct experiments to demonstrate that:
(1) \methodname{} substantially improves training efficiency without compromising performance;
(2) predicted-KL enables stable single-node RL training for more than 1,000 steps; and
(3) ablation analyses provide deeper understanding of our method.

\subsection{General Experimental Setup}

We primarily evaluate Qwen models ranging from 1.5B to 27B parameters, including Qwen2.5-Math-1.5B/7B and Qwen3.8-27B, on mathematical reasoning tasks. We also investigate long-horizon training of the 27B model on the Reasoning-Gym Hard mixture. Unless otherwise specified, we use PPO with verifiable rewards, sampling one response per prompt and using a global running reward baseline for advantage estimation. This deliberately simple setup isolates the effects of our method from potential gains introduced by more sophisticated RL algorithms. By default, all experiments run on a single node equipped with eight H100 80GB GPUs and 128 CPU cores, with four GPUs allocated to FSDP training and the remaining four to independent generation engines.

\begin{figure}[h]
\centering
\includegraphics[width=\linewidth]{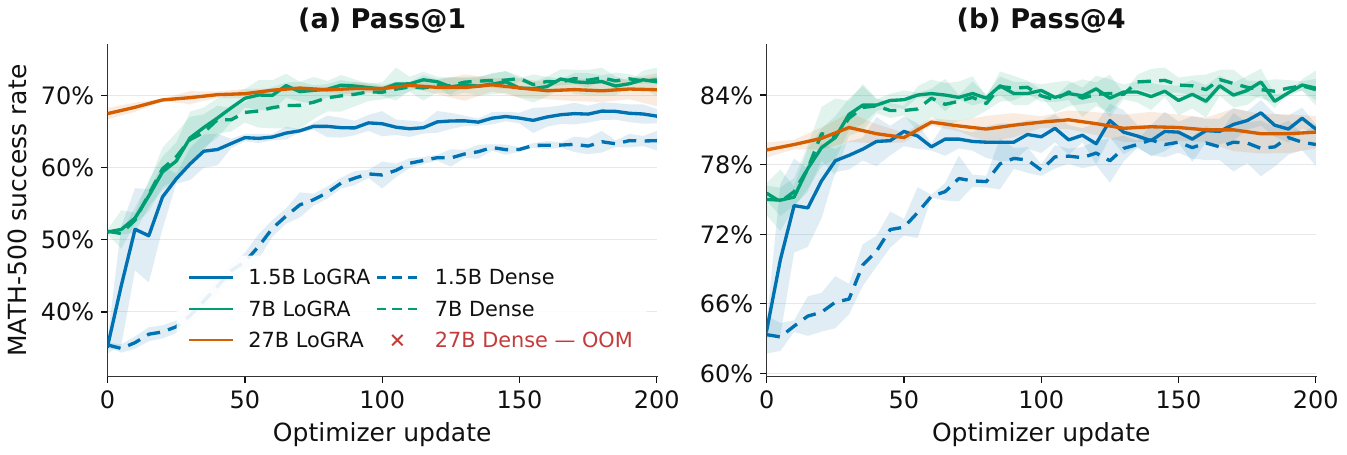}
\caption{\textbf{Learning performance across model scales.} MATH-500 accuracy over the first 200 updates. Curves and bands show means and $\pm$ one sample standard deviation over three seeds. }
\label{fig:flora-training}
\end{figure}

\subsection{Main Experiments}

We primarily compare \methodname{} with dense-gradient training on DAPO-Math-7.5K. We evaluate on MATH-500 every five updates for the 1.5B and 7B models and every ten updates for the 27B model. Each batch contains 128 prompts, with one response per prompt, a 4,096-token context limit, and one policy update using a global reward baseline. \methodname{} uses rank-256 RowAdam with mismatch subtraction and a predicted-KL budget annealed from \(2 \times 10^{-4}\) to \(2 \times 10^{-5}\), whereas the dense baseline uses Adam with learning-rate annealing. We run three runs for 20,000 updates at 1.5B and 7B and 200 updates at 27B, retaining the 300-update annealing schedule for the latter.

\begin{figure}[t]
\centering
\includegraphics[width=\linewidth]{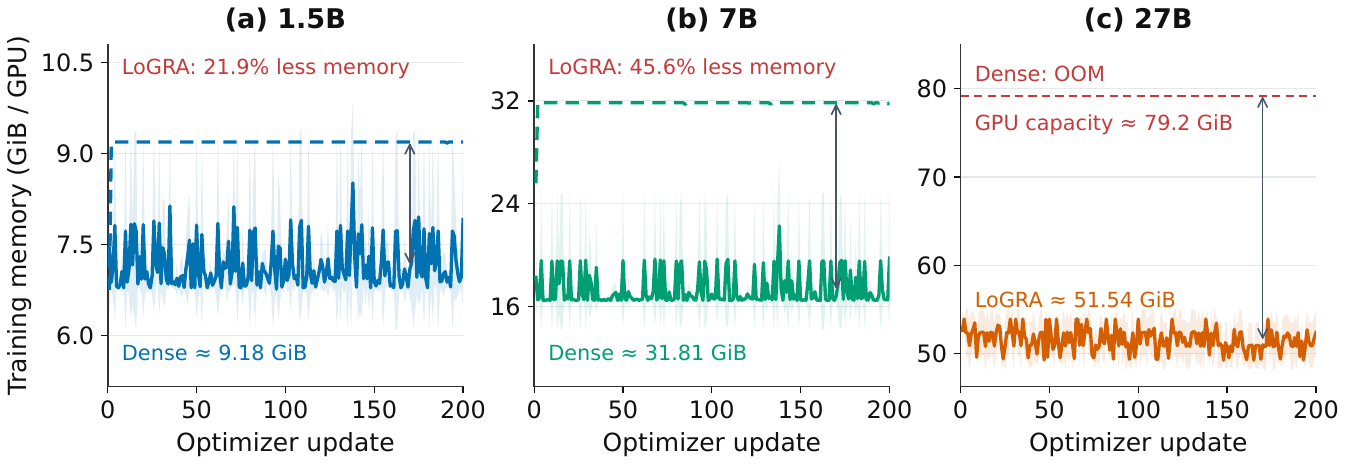}
\caption{\textbf{Training memory across model scales.} Per-update peak allocated memory, averaged over three seeds, with $\pm$ one sample standard deviation. Arrows compare temporal means over updates 1--200 at 1.5B/7B; at 27B, the arrow shows headroom to device capacity. }
\label{fig:training-memory}
\label{fig:controlled-efficiency}
\end{figure}

\begin{table}[htbp]
\centering
\caption{\textbf{Accuracy, memory, and throughput of the main experiments.} Memory and throughput cover full runs; accuracy peaks use updates 0--200. Peak memory is the maximum of the three-seed mean curve. }
\label{tab:plotted-runs-systems}
\small
\setlength{\tabcolsep}{5pt}
\renewcommand{\arraystretch}{1.02}
\resizebox{\linewidth}{!}{%
\begin{tabular}{@{}llcrrrrr@{}}
\toprule
Model & Method & $n$ & \shortstack{Average memory $\downarrow$\\(GiB/GPU)} & \shortstack{Peak memory $\downarrow$\\(GiB/GPU)} & \shortstack{Updates/h $\uparrow$\\\strut} & \shortstack{Peak Pass@1 $\uparrow$\\(\%)} & \shortstack{Peak Pass@4 $\uparrow$\\(\%)} \\
\midrule
1.5B & Dense & 3 & $9.18 \pm 0.00$ & $9.19 \pm 0.00$ & $54.28 \pm 0.40$ & $63.77 \pm 1.35$ & $80.33 \pm 1.47$ \\
\rowcolor{black!8}
1.5B & \methodname{} & 3 & $\mathbf{7.18 \pm 0.10}$ & $\mathbf{8.66 \pm 1.62}$ & $\mathbf{55.44 \pm 0.41}$ & $\mathbf{67.87 \pm 1.23}$ & $\mathbf{82.47 \pm 1.10}$ \\
\addlinespace[1.5pt]
7B & Dense & 3 & $31.82 \pm 0.00$ & $31.85 \pm 0.00$ & $25.28 \pm 0.68$ & $72.48 \pm 0.35$ & $85.33 \pm 0.83$ \\
\rowcolor{black!8}
7B & \methodname{} & 3 & $\mathbf{17.29 \pm 0.34}$ & $\mathbf{22.26 \pm 4.98}$ & $\mathbf{25.73 \pm 0.28}$ & $72.33 \pm 1.33$ & $85.13 \pm 1.14$ \\
\addlinespace[1.5pt]
27B & Dense & 0 & \multicolumn{5}{c}{\textcolor{red!75!black}{OOM}} \\
\rowcolor{black!8}
27B & \methodname{} & 3 & $\mathbf{51.54 \pm 0.17}$ & $\mathbf{53.88 \pm 0.00}$ & $\mathbf{6.99 \pm 0.02}$ & $\mathbf{71.52 \pm 0.94}$ & $\mathbf{81.87 \pm 0.81}$ \\
\bottomrule
\end{tabular}%
}
\end{table}

\paragraph{Learning performance.}
At 1.5B, \methodname{} achieves peak Pass@1 of $67.87\%$, compared with $63.77\%$ for Dense; peak Pass@4 increases from $80.33\%$ to $82.47\%$.
At 7B, the methods attain similar peak accuracy: $72.33\%$ versus $72.48\%$ Pass@1 and $85.13\%$ versus $85.33\%$ Pass@4 for \methodname{} and Dense, respectively.
\methodname{} also trains the 27B model successfully, reaching $71.52\%$ Pass@1 and $81.87\%$ Pass@4, whereas Dense exhausts GPU memory at its first Adam-state allocation.
These results show that compressed updates preserve competitive learning performance in the evaluated configurations while making the largest model trainable under the same GPU budget.

\paragraph{Memory and throughput.}
Figure~\ref{fig:training-memory} shows that \methodname{} lowers average per-update memory at 1.5B and 7B.
Over the complete runs, average memory decreases from 9.18 to 7.18 GiB at 1.5B and from 31.82 to 17.29 GiB at 7B, reductions of 21.8\% and 45.7\%, respectively.
Under the seed-averaged peak definition, \methodname{} reduces peak memory from 9.19 to 8.66 GiB at 1.5B and from 31.85 to 22.26 GiB at 7B.
At 27B, \methodname{} uses 51.54 GiB on average and reaches a peak seed-averaged memory of 53.88 GiB; Dense fails before completing an update, precluding a measured memory-reduction ratio or throughput comparison.
End-to-end update rates are similar where both methods complete training: 55.44 versus 54.28 updates/h at 1.5B and 25.73 versus 25.28 updates/h at 7B for \methodname{} and Dense, respectively. The main systems benefit is reduced memory use and feasibility at 27B, with modest observed throughput gains at 1.5B and 7B.

\subsection{Sustained 27B training on a single eight-GPU node}
\label{sec:sustained-27b}

\begin{figure}[!htb]
  \centering
  \includegraphics[width=\linewidth]{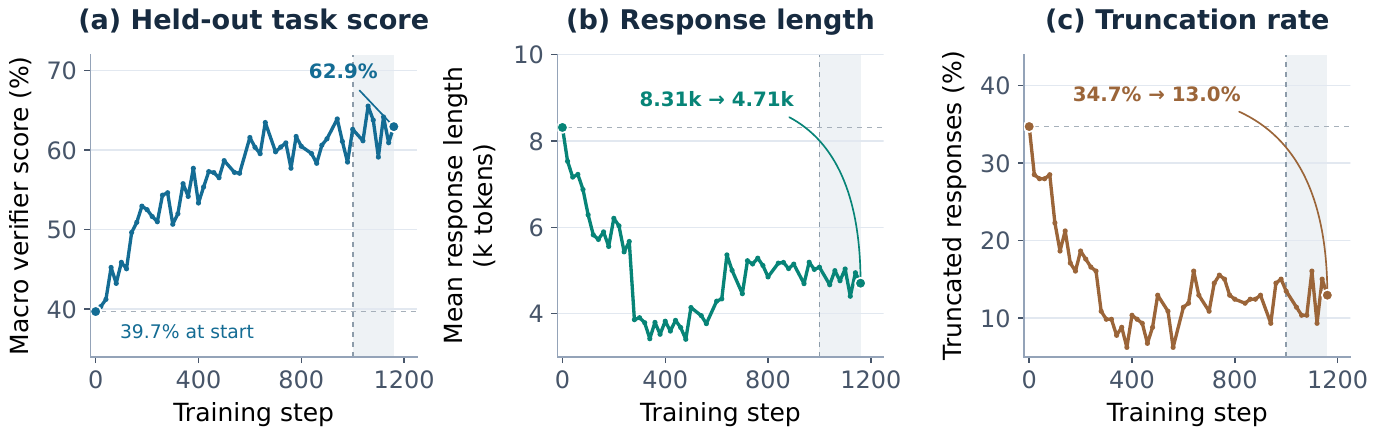}
  \caption{\textbf{Sustained 27B training on a single eight-GPU node.}
  Qwen3.8-27B trained with \methodname{} on the Reasoning-Gym Hard mixture.
  \textbf{(a)} Macro-averaged verifier score across held-out tasks.
  \textbf{(b)} Mean evaluation response length.
  \textbf{(c)} Fraction of evaluation responses truncated at the 16,384-token context limit.
  All panels describe one training trajectory.}
  \label{fig:long27-stability}
\end{figure}

We demonstrate that \methodname{} with predicted-KL enables stable 27B model training for over 1,000 steps on a single node. Specifically, we train Qwen3.8-27B on the Reasoning-Gym Hard mixture. For efficiency purpose, we estimate (q(D)) using a finite-difference probe based on direct KL evaluations, rather than the JVP-based estimator in Equation~\ref{eq:fisher}.
The dataset contains 38,948 training prompts and 193 hard held-out problems across eight tasks. 
We use rank-256 sketches and anneal the predicted-KL budget. Every 20 steps, we evaluate the model using the macro-averaged verifier score and monitor response length and truncation rate. 

The held-out macro score increases from $39.69\%$ initially to $62.94\%$, with a peak of $65.52\%$ at step 1,060 (Figure~\ref{fig:long27-stability}a).
The final five observed evaluations range from $59.11\%$ to $64.16\%$, showing no sustained collapse.
Over the same trajectory, mean response length decreases from 8,312 to 4,705 tokens, while the truncation rate falls from $34.72\%$ to $12.95\%$ (Figure~\ref{fig:long27-stability}b,c).
This trajectory demonstrates that the \methodname{} recipe can sustain learning of a 27B model beyond 1,100 logged training steps on one eight-GPU node, without sustained deterioration in the observed evaluation scores or runaway response-length growth.

\subsection{Further Analysis}

\subsubsection{Design Choices in Low-Rank Gradient Compression}
\label{sec:compression-components}

\begin{figure}[!htb]
\centering
\includegraphics[width=\linewidth]{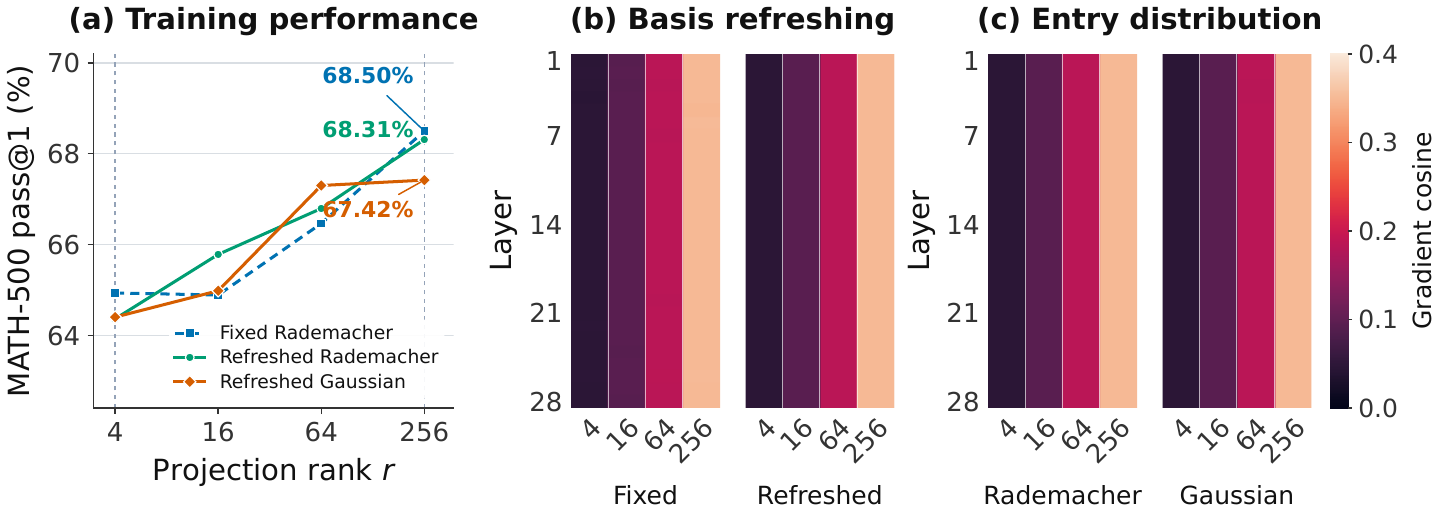}
\caption{\textbf{Components of gradient compression.}
\textbf{(a)} Training performance as a function of projection rank;
\textbf{(b)} Gradient-reconstruction cosine for fixed and refreshed Rademacher bases.
\textbf{(c)} The same diagnostic for refreshed Rademacher and Gaussian bases.
Heatmap rows denote Transformer layers and columns denote ranks, with a shared $[0,0.4]$ color scale; higher values indicate closer alignment with the dense gradient.}
\label{fig:rank-refresh}
\end{figure}

We examine the effects of projection rank, basis refreshing, and projection distribution using Qwen2.5-Math-1.5B on DAPO-Math-7.5k with ranks $r \in {4,16,64,256}$. Fixed Rademacher projections sample matrices once with independent entries $\pm1/\sqrt{r}$, whereas refreshed Rademacher and Gaussian projections resample at every update. Following Flora’s motivation of avoiding confinement to a fixed low-rank subspace~\citep{hao2024flora}, we test whether refreshing benefits RL post-training. Each point represents one run, with MATH-500 pass@1 averaged over eight evaluations between steps 200 and 235. Figure~\ref{fig}(a) shows that rank has the strongest effect: increasing it from 4 to 256 improves pass@1 by 3.02--3.95 percentage points across configurations. At rank 256, fixed Rademacher, refreshed Rademacher, and refreshed Gaussian achieve $68.50\%$, $68.31\%$, and $67.42\%$, respectively. Their ordering varies across ranks, showing no consistent benefit from refreshing, possibly because a fixed low-rank subspace captures sufficient task-relevant signal over this short training horizon.

Figures~\ref{fig:rank-refresh}(b) and~\ref{fig:rank-refresh}(c) evaluate gradient reconstruction quality. We use the initialization and checkpoints at steps 160 and 280 from one fixed-basis rank-64 training trajectory. At each model state, we sample one response for each of 4,096 prompts and divide them into 32 minibatches of 128 responses. All projection conditions use the same responses and dense gradients. Fixed bases are reused across minibatches, whereas refreshed bases are redrawn for each minibatch. For each rank, we evaluate eight projection seeds and compute the Frobenius cosine similarity between the dense gradient $G$ and its reconstruction for all 196 attention and MLP weight matrices. Each heatmap cell is averaged over the seven matrices in each Transformer layer, minibatches, model states, and projection seeds. Figure~\ref{fig:rank-refresh}(b) compares fixed and refreshed Rademacher projections, whereas Figure~\ref{fig:rank-refresh}(c) compares refreshed Rademacher and Gaussian projections.
Reconstruction quality exhibits a strong dependence on rank: the mean cosine similarity increases from approximately $0.046$ at rank 4 to $0.348$ at rank 256. By contrast, the largest difference between corresponding layer--rank cells is only $0.0014$ for fixed versus refreshed Rademacher projections and $0.0008$ for refreshed Rademacher versus Gaussian projections. Although the decrease in mean cosine under Gaussian projections is detectable at ranks 64 and 256, its magnitude remains below $2 \times 10^{-4}$.
Overall, projection rank has a substantially greater effect on single-minibatch gradient reconstruction than either basis refreshing or projection distribution.

\subsubsection{Comparison with LoRA}
\label{sec:lora-comparison}

\begin{table}[htbp]
  \centering
  \caption{\textbf{LoRA and \methodname{} on Qwen2.5-Math-1.5B.} Values are mean $\pm$ sample standard deviation over three seeds.}
  \label{tab:lora-rpga-flora}
  \small
  \setlength{\tabcolsep}{5pt}
  \renewcommand{\arraystretch}{1.12}
  \resizebox{\linewidth}{!}{%
  \begin{tabular}{@{}lrrrrr@{}}
    \toprule
    Method & \shortstack{Average memory $\downarrow$\\(GiB)} & \shortstack{Peak memory $\downarrow$\\(GiB)} & \shortstack{Pass@1 $\uparrow$\\(\%)} & \shortstack{Pass@4 $\uparrow$\\(\%)} & \shortstack{Seconds/update $\downarrow$\\(s)} \\
    \midrule
    LoRA & $13.21 \pm 0.07$ & $13.38 \pm 0.00$ & $70.45 \pm 3.09$ & $81.40 \pm 2.09$ & $10.63 \pm 0.10$ \\
    \rowcolor{black!8}
    \methodname{} & $\mathbf{7.18 \pm 0.10}$ & $\mathbf{8.66 \pm 1.62}$ & $68.47 \pm 0.95$ & $\mathbf{81.80 \pm 1.06}$ & $11.31 \pm 0.09$ \\
    \bottomrule
  \end{tabular}%
  }
\end{table}

LoRA learns low-rank weight adapters, whereas \methodname{} compresses gradients and updates to the model weights. 
We compare their memory use, accuracy, and update time. 
We compare rank-256 configurations of Qwen2.5-Math-1.5B trained on DAPO-Math-7.5k. For this comparison, we use the first 300 training steps of each method. LoRA uses AdamW with a constant learning rate of $2\times10^{-4}$; \methodname{} uses a predicted-KL budget annealed from $2\times10^{-4}$ to $2\times10^{-5}$. 

Average memory is the temporal mean of the three-seed mean per-update memory curve; peak memory is its maximum. The per-step measurements are peak allocated training memory, excluding generation memory. Seconds/update measures the policy-training call, including call overhead but excluding response generation and weight synchronization. 

Table~\ref{tab:lora-rpga-flora} shows an obvious memory advantage for \methodname{} under the recorded recipes, with average and peak memory of 7.18 and 8.66 GiB, compared with 13.21 and 13.38 GiB for LoRA. LoRA attains higher mean Pass@1 and a shorter policy-training call, while \methodname{} attains a 0.40-percentage-point higher mean Pass@4. Thus, these results motivate gradient compression as an alternative when training memory is the limiting resource, but do not establish an accuracy or speed advantage.

\subsubsection{KL Budget Control}

\begin{figure}[htbp]
  \centering
  \includegraphics[width=\linewidth]{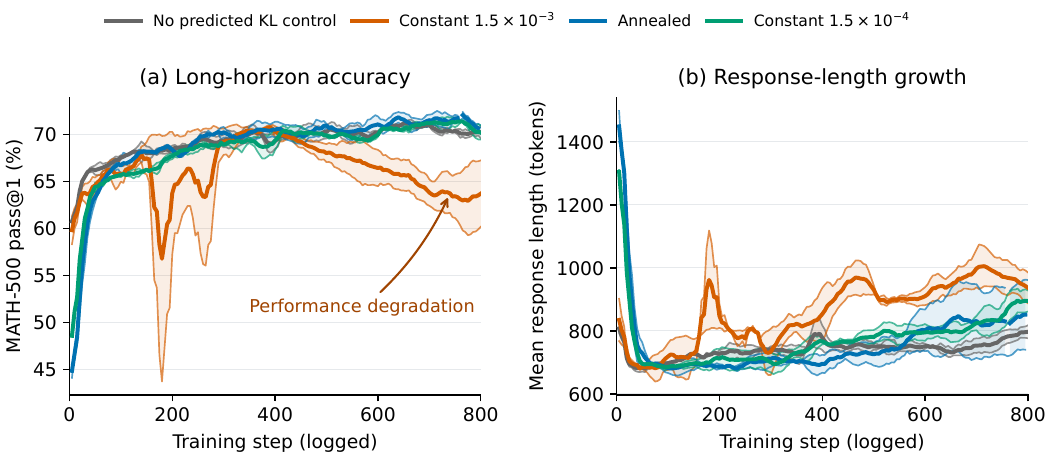}
  \caption{\textbf{Long-horizon accuracy and response length under different KL controls.} Thick solid lines show two-seed means, thin lines individual runs, and shading their range. Curves use five-evaluation moving means.}
  \label{fig:long-horizon-step-control}
\end{figure}
\vspace{-3mm}

We then study whether predicted-KL step control improves long-horizon learning and how constant versus annealed KL budgets affect accuracy and response length. We use Qwen2.5-Math-1.5B on DAPO-Math-7.5k with rank-256 Rademacher projections refreshed every update, SGD, and a global running reward baseline.
We compare a fixed relative update norm of $6\times10^{-5}$ without KL control, constant KL budgets of $1.5\times10^{-3}$ and $1.5\times10^{-4}$, and a cosine-annealed budget from $2\times10^{-4}$ to $2\times10^{-5}$ over 1,500 steps.
Each setting has two seeds; these KL-controlled settings do not subtract mismatch or use realised-KL feedback.
Figure~\ref{fig:long-horizon-step-control} plots MATH-500 pass@1 and response length over the first 800 logged steps, evaluating four responses per problem every five steps.
Averaging raw evaluations over steps 700--800 within each run and then across seeds, pass@1 is $70.30\%$ without KL control, $63.43\%$ with the large constant budget, $71.03\%$ with the small constant budget, and $71.47\%$ with annealing.
Corresponding response lengths are 774, 976, 858, and 832 tokens.

These results support the usefulness of predicted-KL step control with a conservative budget. The small constant budget improves tail accuracy over no control while maintaining stable learning. This is consistent with the trust-region motivation of TRPO~\citep{schulman2015trust}: limiting policy changes helps ensure that improvements predicted by a local objective translate into better performance, whereas overly large updates can undermine this approximation and destabilize training. 
In our experiments, the large constant budget exhibits performance degradation, while the small and annealed budgets remain stable, with annealing achieving the highest mean tail accuracy. The no-control baseline also remains stable, and the separately collected annealed runs differ in their initial budget, so these two-seed results suggest a benefit from well-calibrated KL control without establishing that it is necessary for stability or that annealing alone explains the gain.
\section{Related Works}

Large language models have emerged as a foundation for general-purpose problem solving~\citep{jimenez2024swe,shao2024deepseekmath,xie2024osworld,zhang2024offline,yuan2026osworld2}, with reinforcement learning (RL) becoming a critical post-training stage for advancing their capabilities~\citep{guo2025deepseek,yu2026dapo,hu2025openrlhf}. However, scaling RL is more challenging than scaling SFT, as RL repeatedly alternates among response generation, reward evaluation, and policy optimization, requiring careful coordination of computation and memory across these stages.

Recent work has addressed these challenges through asynchronous execution and more efficient resource management. 
For example, asynchronous RL overlap rollout with policy training, reducing GPU idle time between stages~\citep{xu2026polar,zhang2026prorl}. 
Other works improve resource allocation and scheduling to accommodate the distinct computational demands of generation and training~\citep{mei2025real}. 
Memory-saving techniques further enable larger models to fit within available hardware: sharding distributes training states across GPUs~\citep{zhao2023pytorch}, activation checkpointing reduces memory consumption by recomputing intermediate activations\citep{chen2016training}, and CPU offloading transfers selected training states to host memory~\citep{rajbhandari2020zero}. 
Our work complements them by compressing the gradients and optimization updates themselves, thereby reducing both training memory and communication overhead during policy synchronization.

Another line of work exploits low-rank structure to reduce training memory. LoRA~\citep{hu2021lora} freezes pretrained weights and optimizes low-rank adapters, reducing the memory required for gradients and optimizer states. GaLore~\citep{su2025galore,zhao2024galore} and Flora~\citep{hao2024flora} instead keep the original weights trainable while compactly representing gradients and optimizer states. In contrast, \methodname{} targets the RL training--generation loop: it accumulates low-rank sketches during backpropagation without materializing full gradients for selected weights and reuses the compressed updates for rollout-policy synchronization. It thus extends compression from training-state storage to gradient computation and policy communication.
APOLLO~\citep{MLSYS2025_437bc4cc} uses random projections to estimate channel-wise adaptive scaling factors, reducing optimizer-state memory while retaining full-gradient updates. In comparison, LoGRA directly accumulates gradient sketches and constructs low-rank updates without materializing full gradients for selected weights.

\section{Conclusion}
We presented \methodname{}, a memory-efficient approach to LLM RL post-training that uses low-rank gradient sketches throughout gradient accumulation, weight updates, and policy synchronization. Our results show that compact gradient representations can substantially reduce training memory without sacrificing reasoning performance. In particular, \methodname{} enables sustained training of a 27B model for over 1,100 steps on a single eight-GPU node, where dense Adam exceeds the available memory. Predicted-KL step control complements compression by regulating policy changes, with conservative budgets supporting stable long-horizon learning. Together, these findings highlight gradient representation as a practical lever for lowering the hardware barrier to RL post-training.




\bibliography{references}

@article{sheng2024hybridflow,
  title   = {HybridFlow: A Flexible and Efficient RLHF Framework},
  author  = {Guangming Sheng and Chi Zhang and Zilingfeng Ye and Xibin Wu and Wang Zhang and Ru Zhang and Yanghua Peng and Haibin Lin and Chuan Wu},
  year    = {2024},
  journal = {arXiv preprint arXiv: 2409.19256}
}

@misc{nemo-rl,
title = {NeMo RL: A Scalable and Efficient Post-Training Library},
howpublished = {\url{https://github.com/NVIDIA-NeMo/RL}},
year = {2025},
note = {GitHub repository},
}

@article{xu2026polar,
  title={Polar: Agentic RL on Any Harness at Scale},
  author={Xu, Binfeng and Zhang, Hao and Zhang, Shaokun and Han, Songyang and Liu, Mingjie and Hu, Jian and Diao, Shizhe and Jin, Zhenghui and Zou, Yunheng and Demoret, Michael and others},
  journal={arXiv preprint arXiv:2605.24220},
  year={2026}
}

@inproceedings{zhang2026nemotron,
  title={Nemotron-research-tool-n1: Exploring tool-using language models with reinforced reasoning},
  author={Zhang, Shaokun and Dong, Yi and Zhang, Jieyu and Kautz, Jan and Catanzaro, Bryan and Tao, Andrew and Wu, Qingyun and Yu, Zhiding and Liu, Guilin},
  booktitle={International Conference on Learning Representations},
  volume={2026},
  pages={91437--91453},
  year={2026}
}

@article{bai2023qwen,
  title={Qwen technical report},
  author={Bai, Jinze and Bai, Shuai and Chu, Yunfei and Cui, Zeyu and Dang, Kai and Deng, Xiaodong and Fan, Yang and Ge, Wenbin and Han, Yu and Huang, Fei and others},
  journal={arXiv preprint arXiv:2309.16609},
  year={2023}
}

@article{kingma2014adam,
  title={Adam: A method for stochastic optimization},
  author={Kingma, Diederik P and Ba, Jimmy},
  journal={arXiv preprint arXiv:1412.6980},
  year={2014}
}

@article{zhao2023pytorch,
  title={Pytorch fsdp: experiences on scaling fully sharded data parallel},
  author={Zhao, Yanli and Gu, Andrew and Varma, Rohan and Luo, Liang and Huang, Chien-Chin and Xu, Min and Wright, Less and Shojanazeri, Hamid and Ott, Myle and Shleifer, Sam and others},
  journal={arXiv preprint arXiv:2304.11277},
  year={2023}
}

@article{chen2016training,
  title={Training deep nets with sublinear memory cost},
  author={Chen, Tianqi and Xu, Bing and Zhang, Chiyuan and Guestrin, Carlos},
  journal={arXiv preprint arXiv:1604.06174},
  year={2016}
}

@inproceedings{rajbhandari2020zero,
  title={Zero: Memory optimizations toward training trillion parameter models},
  author={Rajbhandari, Samyam and Rasley, Jeff and Ruwase, Olatunji and He, Yuxiong},
  booktitle={SC20: international conference for high performance computing, networking, storage and analysis},
  pages={1--16},
  year={2020},
  organization={IEEE}
}

@article{hu2021lora,
  title={Lora: Low-rank adaptation of large language models},
  author={Hu, Edward J and Shen, Yelong and Wallis, Phillip and Allen-Zhu, Zeyuan and Li, Yuanzhi and Wang, Shean and Wang, Lu and Chen, Weizhu},
  journal={arXiv preprint arXiv:2106.09685},
  year={2021}
}

@article{zhao2024galore,
  title={Galore: Memory-efficient llm training by gradient low-rank projection},
  author={Zhao, Jiawei and Zhang, Zhenyu and Chen, Beidi and Wang, Zhangyang and Anandkumar, Anima and Tian, Yuandong},
  journal={arXiv preprint arXiv:2403.03507},
  year={2024}
}

@article{su2025galore,
  title={Galore 2: Large-scale llm pre-training by gradient low-rank projection},
  author={Su, DiJia and Gu, Andrew and Xu, Jane and Tian, Yuandong and Zhao, Jiawei},
  journal={arXiv preprint arXiv:2504.20437},
  year={2025}
}

@article{hao2024flora,
  title={Flora: Low-rank adapters are secretly gradient compressors},
  author={Hao, Yongchang and Cao, Yanshuai and Mou, Lili},
  journal={arXiv preprint arXiv:2402.03293},
  year={2024}
}

@article{schulman2025lora,
  author = {John Schulman and Thinking Machines Lab},
  title = {LoRA Without Regret},
  journal = {Thinking Machines Lab: Connectionism},
  year = {2025},
  note = {https://thinkingmachines.ai/blog/lora/},
  doi = {10.64434/tml.20250929},
}

@inproceedings{jimenez2024swe,
  title={Swe-bench: Can language models resolve real-world github issues?},
  author={Jimenez, Carlos E and Yang, John and Wettig, Alexander and Yao, Shunyu and Pei, Kexin and Press, Ofir and Narasimhan, Karthik},
  booktitle={International Conference on Learning Representations},
  volume={2024},
  pages={54107--54157},
  year={2024}
}

@article{shao2024deepseekmath,
  title={Deepseekmath: Pushing the limits of mathematical reasoning in open language models},
  author={Shao, Zhihong and Wang, Peiyi and Zhu, Qihao and Xu, Runxin and Song, Junxiao and Bi, Xiao and Zhang, Haowei and Zhang, Mingchuan and Li, YK and Wu, Yang and others},
  journal={arXiv preprint arXiv:2402.03300},
  year={2024}
}

@article{xie2024osworld,
  title={Osworld: Benchmarking multimodal agents for open-ended tasks in real computer environments},
  author={Xie, Tianbao and Zhang, Danyang and Chen, Jixuan and Li, Xiaochuan and Zhao, Siheng and Cao, Ruisheng and Hua, Toh J and Cheng, Zhoujun and Shin, Dongchan and Lei, Fangyu and others},
  journal={Advances in Neural Information Processing Systems},
  volume={37},
  pages={52040--52094},
  year={2024}
}

@article{zhang2024offline,
  title={Offline training of language model agents with functions as learnable weights},
  author={Zhang, Shaokun and Zhang, Jieyu and Liu, Jiale and Song, Linxin and Wang, Chi and Krishna, Ranjay and Wu, Qingyun},
  journal={arXiv preprint arXiv:2402.11359},
  year={2024}
}

@article{yuan2026osworld2,
  title={OSWorld2. 0: Benchmarking Computer Use Agents on Long-Horizon Real-World Tasks},
  author={Yuan, Mengqi and Zhou, Zilong and Xiong, Xinzhuang and Wu, Weiming and Sun, Jiayang and Song, Jiamin and Cui, Kaiqian and Wang, Bowen and Wu, Haoyuan and Li, Yitong and others},
  journal={arXiv preprint arXiv:2606.29537},
  year={2026}
}

@article{guo2025deepseek,
  title={Deepseek-r1: Incentivizing reasoning capability in llms via reinforcement learning},
  author={Guo, Daya and Yang, Dejian and Zhang, Haowei and Song, Junxiao and Wang, Peiyi and Zhu, Qihao and Xu, Runxin and Zhang, Ruoyu and Ma, Shirong and Bi, Xiao and others},
  journal={arXiv preprint arXiv:2501.12948},
  year={2025}
}

@article{yu2026dapo,
  title={Dapo: An open-source llm reinforcement learning system at scale},
  author={Yu, Qiying and Zhang, Zheng and Zhu, Ruofei and Yuan, Yufeng and Zuo, Xiaochen and Yue, Yu and Dai, Weinan and Fan, Tiantian and Liu, Gaohong and Liu, Lingjun and others},
  journal={Advances in Neural Information Processing Systems},
  volume={38},
  pages={113222--113244},
  year={2026}
}

@inproceedings{hu2025openrlhf,
  title={OpenRLHF: A ray-based easy-to-use, scalable and high-performance rlhf framework},
  author={Hu, Jian and Wu, Xibin and Shen, Wei and Liu, Jason Klein and Wang, Weixun and Jiang, Songlin and Wang, Haoran and Chen, Hao and Chen, Bin and Fang, Wenkai and others},
  booktitle={Proceedings of the 2025 Conference on Empirical Methods in Natural Language Processing: System Demonstrations},
  pages={656--666},
  year={2025}
}

@article{zhang2026prorl,
  title={Prorl agent: Rollout-as-a-service for rl training of multi-turn llm agents},
  author={Zhang, Hao and Liu, Mingjie and Zhang, Shaokun and Han, Songyang and Hu, Jian and Jin, Zhenghui and Zhang, Yuchi and Diao, Shizhe and Lu, Ximing and Xu, Binfeng and others},
  journal={arXiv preprint arXiv:2603.18815},
  year={2026}
}

@article{mei2025real,
  title={Real: Efficient rlhf training of large language models with parameter reallocation},
  author={Mei, Zhiyu and Fu, Wei and Li, Kaiwei and Wang, Guangju and Zhang, Huanchen and Wu, Yi},
  journal={Proceedings of Machine Learning and Systems},
  volume={7},
  year={2025}
}

@article{lightman2023lets,
      title={Let's Verify Step by Step}, 
      author={Lightman, Hunter and Kosaraju, Vineet and Burda, Yura and Edwards, Harri and Baker, Bowen and Lee, Teddy and Leike, Jan and Schulman, John and Sutskever, Ilya and Cobbe, Karl},
      journal={arXiv preprint arXiv:2305.20050},
      year={2023}
}

@inproceedings{schulman2015trust,
  title={Trust region policy optimization},
  author={Schulman, John and Levine, Sergey and Abbeel, Pieter and Jordan, Michael and Moritz, Philipp},
  booktitle={International conference on machine learning},
  pages={1889--1897},
  year={2015},
  organization={Pmlr}
}

@inproceedings{MLSYS2025_437bc4cc,
 author = {Zhu, Hanqing and Zhang, Zhenyu and Cong, Wenyan and Liu, Xi and Park, Sem and Chandra, Vikas and Long, Bo and Pan, David Z. and Wang, Zhangyang and Lee, Jinwon},
 booktitle = {Proceedings of Machine Learning and Systems},
 editor = {M. Zaharia and G. Joshi and Y. Lin},
 pages = {},
 publisher = {MLSys},
 title = {APOLLO: SGD-like Memory, AdamW-level Performance},
 url = {https://proceedings.mlsys.org/paper_files/paper/2025/file/437bc4ccafd3fc6d4289bd10940be42b-Paper-Conference.pdf},
 volume = {7},
 year = {2025}
}
\bibliographystyle{references}

\clearpage
\appendix
\section{Theoretical Derivations}
\label{app:theory}

\subsection{From policy KL to changes in token scores}
\label{app:kl-derivation}
Equation~\ref{eq:quadratic} describes how KL changes with the update scale; Equation~\ref{eq:fisher} gives its coefficient in terms of changes in token scores. We derive them in this order. Throughout, the proposed parameter update $D$ and the nonempty set of contexts $\mathcal P$ are fixed. Each context $s$ consists of a prompt and the response tokens preceding one next-token prediction. Both policies are evaluated on this same text.

\paragraph{Measure the change along an update.}
Let $a$ be a scalar multiplier: $a=0$ leaves the weights unchanged, while $a=1$ applies the full proposal $D$. For vocabulary token $v$, write $p_v(a)=\pi_{\theta+aD}(v\mid s)$ and $p_v=p_v(0)$. We use natural logarithms and the same temperature $T>0$ for both policies. Their KL at context $s$ is
\begin{equation}
 K_s(a)=\sum_v p_v\log\frac{p_v}{p_v(a)}.
 \label{eq:app-kl-context}
\end{equation}
A prime denotes a derivative with respect to the multiplier $a$. Thus $p_v'(0)$ measures how quickly token $v$'s probability initially changes along $D$, whereas $K_s'(0)$ measures the initial slope of the KL. At $a=0$ the two policies coincide, so $K_s(0)=0$. Moreover,
\begin{equation}
 K_s'(0)=-\sum_v p_v'(0)=0,
 \label{eq:app-kl-first}
\end{equation}
because $\sum_v p_v(a)=1$ for every $a$. Individual token probabilities can change to first order, but the KL has zero slope at the point where the distributions agree.

\paragraph{Obtain Equation~\ref{eq:quadratic}: the quadratic dependence on step size.}
The second derivative $K_s''(0)$ measures the local curvature: how quickly the KL's slope changes as we move away from the current policy. Differentiating again gives
\begin{equation}
 K_s''(0)=\sum_v\frac{[p_v'(0)]^2}{p_v}-\sum_v p_v''(0)
 =\sum_v p_v\left[\left.\frac{\mathrm d}{\mathrm da}\log p_v(a)\right|_{a=0}\right]^2.
 \label{eq:app-kl-second}
\end{equation}
The term $\sum_v p_v''(0)$ vanishes because total probability remains one. For the local expansion, assume the logits have continuous, bounded third derivatives along $D$ near $a=0$. Taylor's formula then writes the KL for a small multiplier $\alpha$ as
\begin{equation}
 K_s(\alpha)=K_s(0)+\alpha K_s'(0)+\frac{\alpha^2}{2}K_s''(0)+O(\alpha^3)
 =\frac{\alpha^2}{2}K_s''(0)+O(\alpha^3).
 \label{eq:app-kl-local}
\end{equation}
Here $O(\alpha^3)$ collects terms whose magnitude is bounded by a constant times $|\alpha|^3$ near zero, for fixed $D$. Averaging over the examined predictions gives
\begin{equation}
 \frac{1}{|\mathcal P|}\sum_{s\in\mathcal P}K_s(\alpha)
 =\alpha^2q(D)+O(\alpha^3),\qquad
 q(D)=\frac{1}{2|\mathcal P|}\sum_{s\in\mathcal P}K_s''(0).
 \label{eq:app-kl-average}
\end{equation}
Dropping the higher-order remainder gives Equation~\ref{eq:quadratic}. Thus $q(D)$ is half the average KL curvature along the proposal: a larger value means that the same update scale changes the policy more. Within this approximation, halving the scale reduces KL by a factor of four.

\paragraph{Obtain Equation~\ref{eq:fisher}: express the curvature using token scores.}
Let $z_v(a;s)$ be the logit, or score, of token $v$ at context $s$ under weights $\theta+aD$. Define $\dot z_v(s)=\left.\mathrm d z_v(a;s)/\mathrm da\right|_{a=0}$, its initial rate of change along the proposal. Using
\[
 p_v(a)=\frac{\exp(z_v(a;s)/T)}{\sum_u\exp(z_u(a;s)/T)},
\]
where $u$ also ranges over vocabulary tokens, differentiating the log probability yields
\begin{equation}
 \left.\frac{\mathrm d}{\mathrm da}\log p_v(a)\right|_{a=0}
 =\frac{1}{T}\left(\dot z_v(s)-\sum_u p_u\dot z_u(s)\right).
 \label{eq:app-logprob-direction}
\end{equation}
The subtraction reflects that probabilities depend on relative scores: a token gains probability when its score increases faster than the probability-weighted average score. Substituting into Equation~\ref{eq:app-kl-second} gives
\[
 K_s''(0)=\frac{1}{T^2}\sum_v p_v\left(\dot z_v(s)-\sum_u p_u\dot z_u(s)\right)^2
 =\frac{1}{T^2}\operatorname{Var}_{v\sim p(s)}[\dot z_v(s)],
\]
where $p(s)$ collects the current token probabilities. This variance measures how differently the scores change, weighted by those probabilities. In particular, adding the same amount to every score leaves softmax probabilities unchanged; correspondingly, identical score-change rates give zero curvature. Substituting this expression into the definition of $q(D)$ above gives exactly Equation~\ref{eq:fisher}:
\[
 q(D)=\frac{1}{2|\mathcal P|T^2}\sum_{s\in\mathcal P}
 \operatorname{Var}_{v\sim p(s)}[\dot z_v(s)].
\]
The average weights examined predictions equally. The probe evaluates the variance over the full vocabulary at each selected prediction; only the contexts are subsampled. It computes score-change rates before applying the update, using the factors $D_W=-\eta UA$ for each selected weight matrix. Neither a stored dense gradient nor a dense curvature matrix is needed.

\subsection{Scaling an update to a KL budget}
\label{app:kl-scaling}
The multiplier $\alpha$ controls how much of the proposed update $D$ we apply: $\alpha=1$ keeps it unchanged, while $\alpha=1/2$ halves it. Let $q(D)$ be the predicted KL for the full proposal and let $\delta>0$ be the KL budget, the allowed average predicted change in the policy. We choose $\alpha$ by comparing these two quantities. If the proposal exceeds the budget, we shrink it; if it falls below the budget, we enlarge it.

The key is that predicted KL scales with the \emph{square} of the update multiplier, as derived in Appendix~\ref{app:kl-derivation}. To match the budget, we therefore solve $\alpha^2q(D)=\delta$, giving $\alpha=\sqrt{\delta/q(D)}$ for finite positive $q(D)$. For example, if the proposal predicts four times the allowed KL, halving the update brings its predicted KL down to the budget. Conversely, if it predicts one quarter of the budget, doubling the update would reach the budget. We cap this enlargement at a chosen maximum multiplier $\alpha_{\max}>0$, giving Equation~\ref{eq:alpha}:
\begin{equation}
 \alpha=\min\!\left\{\alpha_{\max},\sqrt{\delta/q(D)}\right\}.
 \label{eq:app-budget-scale}
\end{equation}
The cap prevents a very small KL prediction from producing an excessively large update. This rule changes the update's size, not its direction.

\section{RowAdam}
\label{app:rowadam}

In the main experiments, we use RowAdam to make the comparison with dense Adam by retaining adaptive scaling based on past gradients. 
RowAdam is not identical to Adam: it adapts the scale of each row rather than each individual parameter, and does not maintain first-moment momentum.

Let $S_t$ be the normalized gradient sketch at update $t$. For row $i$, RowAdam maintains a running estimate $v_{t,i}$ of its squared magnitude and uses it to scale the row:
\begin{equation}
 \begin{aligned}
 v_{t,i}&=\beta_2v_{t-1,i}+(1-\beta_2)\frac{1}{r}\sum_{j=1}^{r}(S_t)_{ij}^{2},\\
 (\widetilde U_t)_{i,:}&=\frac{(S_t)_{i,:}}{\sqrt{\widehat v_{t,i}}+\epsilon_t},
 \qquad \widehat v_{t,i}=\frac{v_{t,i}}{1-\beta_2^t}.
 \end{aligned}
 \label{eq:app-row-moment}
\end{equation}
We initialize $v_{0,i}=0$; the correction in $\widehat v_{t,i}$ accounts for this zero initialization. The parameter $\beta_2$ controls how much history is retained. Rows with consistently larger squared gradients receive smaller multipliers. After this row scaling, we rescale $\widetilde U_t$ to match the original sketch's Frobenius norm, yielding the factor $U_t$ used for the weight update. Thus RowAdam adjusts the relative contribution of different rows while preserving the sketch's overall magnitude.
We use $\beta_2=0.95$. The stabilizer $\epsilon_t$ is $10^{-8}$ times the mean of $\sqrt{\widehat v_{t,i}}$ across rows, with a positive numerical floor; zero sketches produce zero updates. The row statistics and their update counter persist across projection refreshes. Each selected matrix requires only $d$ second-moment values, rather than one value per weight. 

\section{Data Details}
\label{app:data}

\subsection{Mathematical reasoning}
We construct DAPO-Math-7.5K from the DAPO-Math-17k training split, using the corresponding release on Hugging Face~\citep{yu2026dapo}. The preparation script streams the source in its dataset order, retains the first occurrence of each prompt, and stops after collecting 7,500 unique prompts. The deduplication key is the content of the first prompt message. 

Evaluation uses all 500 problems in the test split of MATH-500~\citep{lightman2023lets}. Each problem is converted to a user message, and its reference answer is stored as a string for the verifier. Training and evaluation use the model's chat template. The training-subset deduplication described above operates within the training source; it is not a cross-dataset semantic decontamination procedure.

Evaluation generates four responses per problem at temperature $0.7$ and top-$p$ $0.7$. Pass@1 is the mean binary correctness over these 2,000 responses. The verifier extracts and checks the final mathematical answer; an unparseable answer receives no correctness credit. Training generates one response per prompt at temperature $1.0$ and top-$p$ $1.0$, with a 4,096-token total context limit.

\subsection{Reasoning-Gym Hard mixture}

\begin{table}[htbp]
\centering
\caption{Reasoning-Gym Hard mixture.}
\label{tab:app-rg-data}
\small
\begin{tabular}{lrr}
\toprule
Task & Training problems & Evaluation problems \\
\midrule
\texttt{codeio} & 4,868 & 25 \\
\texttt{emoji\_mystery} & 4,752 & 22 \\
\texttt{rectangle\_count} & 5,000 & 25 \\
\texttt{word\_ladder} & 4,994 & 25 \\
\texttt{pool\_matrix} & 5,000 & 25 \\
\texttt{tsumego} & 5,000 & 25 \\
\texttt{sentence\_\allowbreak reordering} & 4,334 & 21 \\
\texttt{manipulate\_matrix} & 5,000 & 25 \\
\midrule
Total & 38,948 & 193 \\
\bottomrule
\end{tabular}
\end{table}

The long-horizon 27B experiment uses an eight-task Reasoning-Gym mixture with task-specific verifiers.  We select tasks with mean verifier scores of approximately $0.25$--$0.60$ in a preliminary evaluation of 89 tasks, excluding those whose failures mainly reflect response truncation or formatting errors. This provides a challenging training mixture with room for improvement.

For each task, we generate 5,000 candidate training problems and 25 evaluation problems at the default difficulty. 
We remove duplicate training questions and exclude training--evaluation overlaps; Table~\ref{tab:app-rg-data} reports the final counts. 
Each task's verifier returns a score in ([0,1]). We report the equally weighted mean of the eight task-average scores. 
We report the equally weighted mean of the eight task-average scores. Evaluation uses one response per problem, temperature $0.7$, top-$p$ $0.95$, and a 16,384-token context limit. 

\section{Hyperparameters}
\label{app:hyperparameters}

\subsection{Main mathematical-reasoning experiments}
Table~\ref{tab:app-main-settings} summarizes the settings for Table~\ref{tab:plotted-runs-systems}. Each run uses one node with eight H100 80GB GPUs and 128 CPU cores: four FSDP training GPUs and four independent generation engines (tensor-parallel degree one). Both use BF16 computation, activation checkpointing, no CPU offloading, and full-weight policy synchronization.

\begin{table}[htbp]
\centering
\caption{Main-experiment settings. Schedule horizons count actual optimizer updates.}
\label{tab:app-main-settings}
\small
\setlength{\tabcolsep}{5pt}
\begin{tabular}{p{.31\linewidth}p{.28\linewidth}p{.30\linewidth}}
\toprule
Setting & Dense & \methodname{} \\
\midrule
Training seeds & $1,2,42$ & $1,2,42$ \\
Prompts / responses per update & $128$ / $128$ & $128$ / $128$ \\
Microbatch per training GPU & $1$ & $1$ \\
Optimizer & Adam (zero weight decay) & RowAdam \\
Moment decay & $(\beta_1,\beta_2)=(0.9,0.95)$ & $\beta_2=0.95$; no first moment \\
Epsilon & $10^{-8}$, absolute & $10^{-8}$, relative \\
Projection & Not applicable & Rank $256$, Rademacher \\
Projection refresh & Not applicable & Every update \\
Step-size rule & LR $10^{-6}\rightarrow10^{-7}$ & Predicted KL $2\!\times\!10^{-4}\rightarrow2\!\times\!10^{-5}$ \\
Schedule & $9$-update warmup; cosine decay & Cosine budget decay \\
Mismatch subtraction & Not applicable & Enabled; budget floor $50\%$ \\
Evaluation interval  & $5$ / $5$ / OOM & $5$ / $5$ / $10$ \\
\bottomrule
\end{tabular}
\end{table}

\paragraph{KL budget.}
For update $u=1,\ldots,H$, the main experiments use
\begin{equation}
 \delta_u=\delta_{\min}+\frac{\delta_{\max}-\delta_{\min}}{2}
 \left[1+\cos\left(\pi\frac{u-1}{H-1}\right)\right],
 \label{eq:app-budget-schedule}
\end{equation}
with $H=300$, $\delta_{\max}=2\times10^{-4}$, and $\delta_{\min}=2\times10^{-5}$. The 27B runs stop at 200 updates without shortening the schedule.

Mismatch subtraction accounts for differences between rollout-engine and trainer probabilities at the same nominal weights. We average rollout-minus-trainer log probabilities over generated tokens in completed, nontruncated responses, clamp the estimate below at zero, and smooth it with moving-average weight $0.3$. Denoting this estimate by $\kappa_u$, the effective budget is
\begin{equation}
 \delta_u^{\mathrm{eff}}=\max(\delta_u-\kappa_u,\,0.5\delta_u).
 \label{eq:app-effective-budget}
\end{equation}
Without a valid measurement, no subtraction is applied. This adjustment is a budgeting heuristic, not an exact decomposition of KL.

\subsection{Long-horizon 27B training}
The Reasoning-Gym run in Figure~\ref{fig:long27-stability} uses seed $42$, 128 prompts with one response each, microbatch size one, and a 16,384-token context limit. Training temperature and top-$p$ are both $1.0$. It uses rank-256 refreshed projections, SGD sketch updates, a global running baseline, and a KL budget annealed from $2\times10^{-4}$ to $2\times10^{-5}$ over 1,000 merges then held constant. Four training GPUs and four generation engines share one node.
Evaluation is scheduled every 20 logged steps; 

\subsection{Reconstruction heatmaps.}

We use initialization and steps 160 and 280 of one fixed-basis rank-64 trajectory. At each state, all conditions share 4,096 responses, grouped into 32 batches of 128, and their dense gradients. We measure reconstruction before optimizer transformations:
\[
 \widehat G=(GA^\top)A,\qquad
 \operatorname{cos}(G,\widehat G)=
 \frac{\sum_{i,j}G_{ij}\widehat G_{ij}}{\|G\|_F\|\widehat G\|_F}.
\]
Each heatmap cell equally averages the seven attention/MLP matrices in a layer, 32 batches, three model states, and eight random projection realizations; 28 layers contain 196 matrices in total. Fixed bases are reused across batches; refreshed bases are redrawn.

\subsection{LoRA}
Table~\ref{tab:lora-rpga-flora} uses rank 256 and Qwen2.5-Math-1.5B, with three runs per method. LoRA uses AdamW with constant learning rate $2\times10^{-4}$, moment decay $(0.9,0.95)$, epsilon $10^{-8}$, and zero weight decay. Both methods use 128 prompts with one response each, microbatch size one, and a 4,096-token context limit; \methodname{} follows Table~\ref{tab:app-main-settings}. LoRA statistics cover 300 logged steps, excluding records without an update from memory/time averages; \methodname{} covers 300 actual updates. The different counters, optimizers, and schedules prevent isolating adapter parameterization alone.
\end{document}